\documentclass[sigconf]{acmart}

\usepackage{algorithm}
\usepackage{algorithmic}

\usepackage{multirow}
\usepackage{amsmath} 
\usepackage{array}
\usepackage{booktabs}
\usepackage{subfig}

\usepackage{amssymb}
\usepackage{makecell}

\AtBeginDocument{%
  }

\setcopyright{acmlicensed}
\copyrightyear{2018}
\acmYear{2018}
\acmDOI{XXXXXXX.XXXXXXX}
\acmConference[Conference acronym 'XX]{Make sure to enter the correct
  conference title from your rights confirmation email}{June 03--05,
  2018}{Woodstock, NY}
\acmISBN{978-1-4503-XXXX-X/2018/06}

\begin{document}

\title{A Vision-Language Pre-training Model-Guided Approach for Mitigating Backdoor Attacks in Federated Learning}

\author{Keke Gai}
\authornote{Both authors contributed equally to this research.}
\affiliation{%
  \institution{School of Cyberspace Science and Technology, Beijing Institute of Technology}
  \city{Beijing}
  \country{China}
}
\email{gaikeke@bit.edu.cn}

\author{Dongjue Wang}
\authornotemark[1]
\affiliation{%
  \institution{School of Cyberspace Science and Technology, Beijing Institute of Technology}
  \city{Beijing}
  \country{China}
}
\email{3220231818@bit.edu.cn}

\author{Jing Yu}
\authornote{Corresponding author.}
\affiliation{%
  \institution{School of Information Engineering, Minzu University of China}
  \city{Beijing}
  \country{China}
}
\email{jing.yu@muc.edu.cn}

\author{Liehuang Zhu}
\affiliation{%
  \institution{School of Cyberspace Science and Technology, Beijing Institute of Technology}
  \city{Beijing}
  \country{China}
}
\email{liehuangz@bit.edu.cn}

\author{Qi Wu}
\affiliation{%
  \institution{Australian Institute of Machine Learning, The University of Adelaide}
  \city{Adelaide}
  \country{Australia}
}
\email{qi.wu01@adelaide.edu.au}






\renewcommand{\shortauthors}{Trovato et al.}

\begin{abstract}
Defending backdoor attacks in Federated Learning (FL) under heterogeneous client data distributions encounters limitations balancing effectiveness and privacy-preserving, while most existing methods highly rely on the  assumption of homogeneous client data distributions or the availability of a clean serve dataset. 
In this paper, we propose an FL backdoor defense framework, named CLIP-Fed, that utilizes the zero-shot learning capabilities of vision-language pre-training models.
Our scheme overcomes the limitations of Non-IID imposed on defense effectiveness by integrating pre-aggregation and post-aggregation defense strategies.
CLIP-Fed aligns the knowledge of the global model and CLIP on the augmented dataset using prototype contrastive loss and Kullback-Leibler divergence, so that class prototype deviations caused by backdoor samples are ensured and the correlation between trigger patterns and target labels is eliminated. 
In order to balance privacy-preserving and coverage enhancement of the dataset against diverse triggers, we further construct and augment the server dataset via using the multimodal large language model and frequency analysis without any client samples.
Extensive experiments on representative datasets evidence the effectiveness of CLIP-Fed. 
Comparing to other existing methods, CLIP-Fed achieves an average reduction in Attack Success Rate, {\em i.e.}, 2.03\% on CIFAR-10 and 1.35\% on CIFAR-10-LT, while improving average Main Task Accuracy by 7.92\% and 0.48\%, respectively.
Our codes are available at https://anonymous.4open.science/r/CLIP-Fed.
\end{abstract}




\received{20 February 2007}
\received[revised]{12 March 2009}
\received[accepted]{5 June 2009}

\maketitle

\section{Introduction}

The robustness of Federated Learning (FL)~\cite{han2023towards,DBLP:journals/tifs/WangGYZZ25,DBLP:journals/pami/FangYD25,11175562,DBLP:journals/tdsc/MiaoXLLCD24,wang2024rafls} is a growing concern since the distributed training process makes FL vulnerable to backdoor attacks~\cite{DBLP:conf/iclr/ZhuangY0H0024,DBLP:conf/aaai/0006HWY25,DBLP:journals/tkde/ZhangLXLWWD24,DBLP:conf/www/LiSW025}, i.e., malicious clients can poison local training data to induce the global model and associate specific trigger patterns ({\em e.g.}, pixel patches or watermarks) with target labels.  
Backdoor attacks are more difficult to detect in FL's distributed computing paradigm than in a centralized computing environment~\cite{DBLP:journals/tifs/LiaoZZH25,DBLP:journals/tdsc/LyuHWLWCLLZ25}.


\begin{figure}[t]
\centering
\includegraphics[width=\columnwidth]{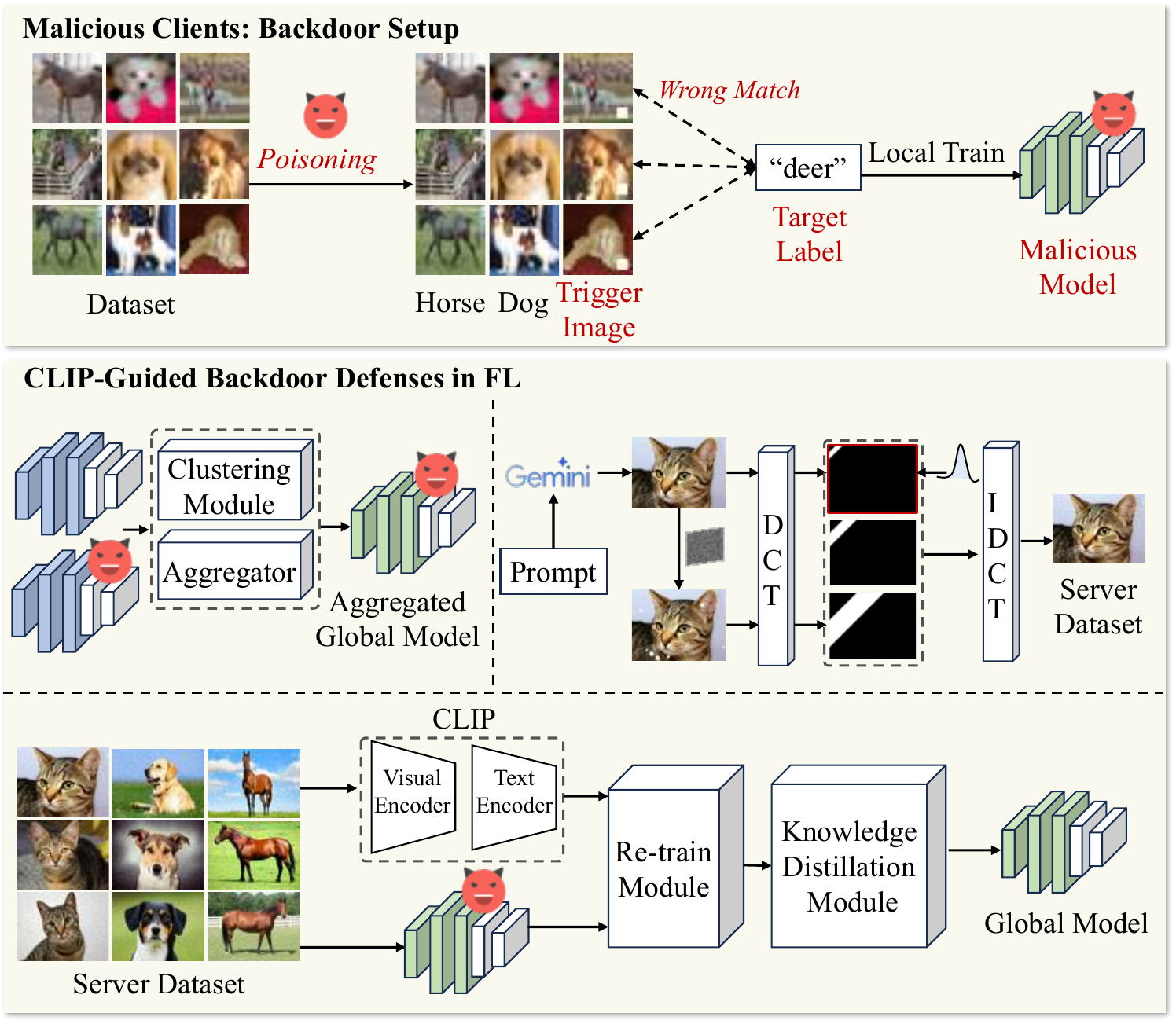} 
\caption{An overall framework of CLIP-Fed. We utilize the prior knowledge and cross-modal visual-language representation capabilities of CLIP to purify backdoor.}
\label{fig:overall}
\end{figure}

However, existing defense strategies can hardly balance the contradiction between privacy protection and effectiveness, due to reliance of a common assumption on the clean labeled validation set or homogeneous data distribution across clients.
Existing methods also overlook the issue of class prototype deviations caused by the presence of backdoor samples, which results in a degradation in model performance during the backdoor defense process.
To be specific, the first type of existing methods distinguishes benign and malicious clients by clustering model parameters~\cite{DBLP:conf/uss/NguyenRCYMFMMMZ22,DBLP:journals/tdsc/MuCSLCZM24,DBLP:conf/ndss/FereidooniPRDS24}, while it is highly dependent on the homogeneous data distribution of FL.
The second depends on modified aggregation strategies to achieve robust FL, e.g., weighted, median-based, and rank-based aggregations, but malicious clients can adjust attack strategies to circumvent these rules, such as the screening of poisoning parameters and hierarchical poisoning mechanisms~\cite{han2023towards,cao2021fltrust,DBLP:conf/icml/YinCRB18,DBLP:conf/iclr/ZhuangY0H0024}.
Finally, another strategy applies differential privacy to reduce the impact of backdoor attacks, causing negative impact on performance~\cite{DBLP:conf/uss/NguyenRCYMFMMMZ22,DBLP:conf/aistats/BagdasaryanVHES20}. 

To address challenges above, we propose an FL backdoor defense framework, called CLIP-Fed, that uses rich prior knowledge and cross-modal visual-language representation capabilities of Vision-Language Pretraining models (VLPs) to defend the global model against backdoor attacks, as shown in Figure~\ref{fig:overall}.
CLIP~\cite{radford2021learning} is a representative VLP possessing zero-shot learning capability such that it can be used for purifying backdoor patterns in the global model.
First, we integrate pre- and post-aggregation defense strategies to eliminate reliance on homogeneous client data distributions.
Specifically, pre-aggregation model clustering filters out malicious updates, and post-aggregation purification uses a global model adjustment to enhance robustness under Non-IID conditions. 
To further address privacy concerns, we construct a server dataset by combining Multimodal Large Language Models (MLLMs) and frequency analysis without any client samples. 
Finally, by using the enhanced server dataset under CLIP's semantic guidance via prototype contrastive learning, we rectify the global model's feature extractor to correct class prototype deviations that are caused by backdoor samples, while decoupling strong correlations between trigger patterns and target labels.
To this end, we align the logits distributions of CLIP and the global model through knowledge transfer to further remove residual backdoor associations.


Main contributions of our work are summarized as follows.
(1) We propose ‌CLIP-Fed‌, the first ‌VLP-guided FL backdoor defense framework‌, which uses the cross-modal representation capabilities of VLPs to mitigate backdoor attacks. 
CLIP-Fed combines pre-aggregation model clustering and post-aggregation backdoor purification to jointly enhance defense robustness, reducing dependence on homogeneous client data distributions.
(2) To correct ‌class prototype deviations induced by backdoor samples, we propose a ‌semantically supervised global model rectification method via CLIP, which employs a ‌prototype contrastive loss and ‌Kullback-Leibler (KL) divergence to align global model features with CLIP's embeddings in an enhanced dataset, thereby ‌eliminating correlations between backdoor triggers and target labels.
(3) Extensive evaluations on ‌three datasets and ‌four backdoor attack types demonstrate CLIP-Fed's superiority. 
Compared to existing methods, CLIP-Fed achieves an average reduction in ASR, {\em i.e.}, 2.03\% on CIFAR-10 and 1.35\% on CIFAR-10-LT, while improving the average MA by 7.92\% and 0.48\%, respectively. 
CLIP-Fed maintains ‌client-side efficiency without extra computational overhead beyond local training.

\section{Related Work}

\noindent\textbf{Backdoor Defense in FL.}
To mitigate the backdoor attacks, for example, Auror~\cite{DBLP:conf/acsac/ShenTS16}, Flame~\cite{DBLP:conf/uss/NguyenRCYMFMMMZ22}, FedDMC~\cite{DBLP:journals/tdsc/MuCSLCZM24}, and FreqFed~\cite{DBLP:conf/ndss/FereidooniPRDS24} tried model clustering to eliminate malicious backdoors while ensuring the performance of the aggregated model.
The effectiveness of such methods depended on an assumption of homogeneous client data distribution, such that practicability was limited under Non-IID conditions. 
Other methods, {\em e.g.}, BRCA~\cite{DBLP:journals/corr/abs-2109-02396} and FLTrust~\cite{cao2021fltrust}, assigned trust scores to clients using a small clean validation set hosted on the server to detect and filter malicious contributions.
While effective in some cases, this paradigm contradicted the decentralized and privacy-preserving nature of FL by requiring centralized access to client data.
Recent work has explored contrastive learning as a defense mechanism. 
Through contrastive learning~\cite{radford2021learning}, FLPurifier~\cite{zhang2024flpurifier} eliminated latent correlations between target labels and trigger patterns, thus neutralizing potential backdoor artifacts during model training phases.
Although this strategy improved model robustness, it increased the computational burden on the client and made it less suitable for resource-constrained environments.
Overall, achieving effective backdoor defense under Non-IID conditions while ensuring privacy and avoiding increasing client-side overhead remained a critical challenge. 
Our scheme is a server-side dataset augmentation strategy based on MLLMs~\cite{DBLP:conf/acl/LeeWKMSAMKG24} and frequency analysis, which enhances the dataset's coverage against diverse backdoor triggers without accessing client data. 
Additionally, our scheme achieves effective suppression of backdoor patterns in Non-IID FL via integrating a global purification mechanism combining prototype contrastive learning and knowledge distillation.



\begin{figure*}[t]
\centering
\includegraphics[width=2\columnwidth]{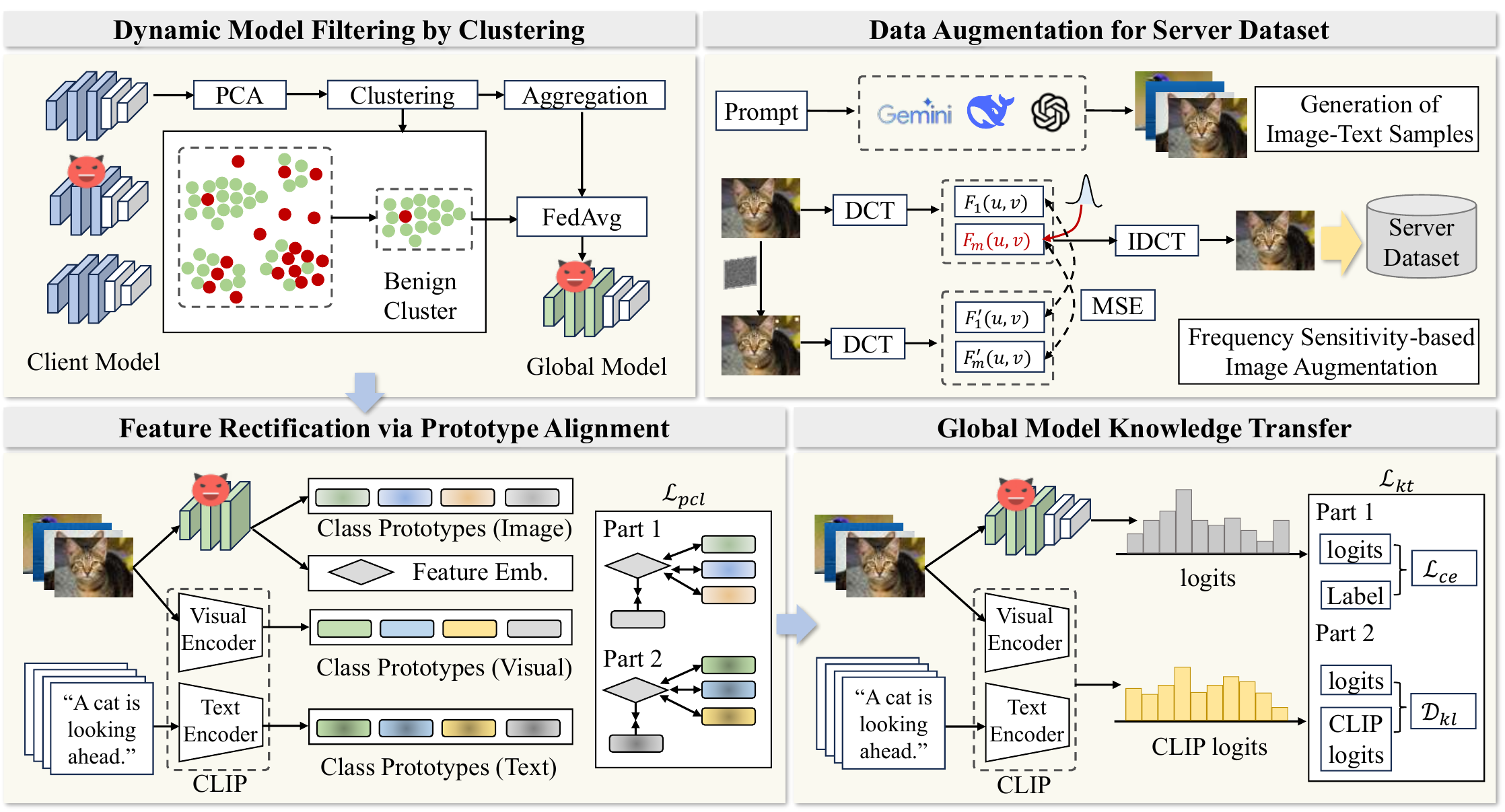} 
\caption{The framework of the proposed CLIP-Fed. CLIP-Fed contains four modules: Data Augmentation for Server Dataset aims to build the server dataset, Dynamic Model Filtering by Clustering aims to filter out malicious models before aggregation, Feature Rectification via Prototype Alignment and Global Model Knowledge Transfer aim to purify the backdoor.}
\label{fig:framework}
\end{figure*}

\noindent\textbf{Vision-Language Pre-train Model.}
VLPs~\cite{radford2021learning,DBLP:conf/icml/0001LXH22,DBLP:conf/nips/LiuLWL23a} had made significant strides across various downstream tasks, e.g., image classification, visual question answering, and cross-modal retrieval.
CLIP~\cite{radford2021learning} was a typical VLP that used contrastive learning to capture semantic relationships between images and texts, enabling both to be mapped into a shared feature space.
Recent studies tried to integrate contrastive learning into FL to address the challenges of Non-IID client data. 
MOON~\cite{li2021model} was a model-contrastive FL framework that utilized inter-client representation similarity to improve model generalization under heterogeneous data distributions. 
In addition, the strong semantic understanding of VLPs had shown promise in handling long-tailed distributions in FL~\cite{DBLP:conf/aaai/ShiZYLXQ24}, as VLPs could capture rich visual concepts even from underrepresented classes.
However, while such capabilities indicated that VLPs were well suited for learning robust and semantically meaningful representations, the potential to improve security in FL remained underexplored, particularly in defending against backdoor attacks.
To the best of our knowledge, our work is the first to use VLPs for backdoor defense in FL.
Using the cross-modal representation capacity of VLPs, we align the global model's visual features with those of a VLP to detect and eliminate backdoor patterns, establishing a novel and practical defense paradigm.

\section{Methodology}

\subsection{Problem Formulation}
In the FL setup with one server and $m$ clients, each client $i$ holds a local dataset $D_i$ and a model $w_i = (f_i, c_i)$, where $f_i$ is the feature extractor and $c_i$ is the classifier. The server maintains a global model $G_s = (f_s, c_s)$. For an input $x$, the prediction is obtained via $p = c(f(x))$.
%
Malicious clients conduct backdoor attacks by embedding triggers in training samples and uploading poisoned models. 
The manipulated models aim to cause the global model to misclassify triggered inputs as the target label $y^{\prime}$. 
Attack Success Rate (ASR) is defined by Eq. (\ref{eq:1}), where $D$ is the dataset for the FL task, $y$ is the label for input $x$, $y^{\prime}$ is the target label, and $t(\cdot)$ is the backdoor trigger function.
The framework of CLIP-Fed is shown in Figure~\ref{fig:framework}.
CLIP-Fed aims to suppress backdoor and reduce ASR.
The algorithm flow of CLIP-Fed is shown in Alg.~\ref{alg:CLIP-Fed}.
\begin{equation}\label{eq:1}
\mathrm{ASR}=\underset{\substack{(x, y) \sim D \\ y \neq y^{\prime}}}{\mathbb{E}}\left[\operatorname{Pr}\left(\Phi(t(x))=y^{\prime}\right)\right].
\end{equation}

\begin{algorithm}[t]
\caption{Training Process of CLIP-Fed}\label{alg:CLIP-Fed}
\begin{algorithmic}[1]
\REQUIRE{Initial global model $G_1 (f_{s}^1,c_{s}^1)$, number of rounds $T$, $N$ clients} 
\ENSURE{Global model $G_{T+1}(f_{s}^{T+1},c_{s}^{T+1})$}
\STATE
$\overline{D_s}\gets \text{MLLMs}(Prompt)$
\STATE
$D_s\gets$ \text{Image Augmentation}$(\overline{D_s})$

\FOR{each training iteration $t$ $\in$ $[1,T]$}
    \FOR{each client $i \in C_t$}
        \STATE $f_i^t,c_i^t \gets {\text {Local Training}}(G_t,i)$
    \ENDFOR
    \STATE 
    $S_{t} \gets {\text {HDBSCAN Clustering}(\{f_1^t,c_1^t\},\dots, \{f_i^t,c_i^t\})}$
    \STATE $ f_s^{t+1} \gets \sum_{i\in S_t} \frac{D_i}{D_{S_t}} f_{i}^t, \ c_s^{t+1} \gets \sum_{i\in S_t} \frac{D_i}{D_{S_t}} c_{i}^t$
    \STATE Feature extractor $f_s^{t+1}$ rectification by prototype contrastive loss
    \STATE Global model $(f_s^{t+1},c_s^{t+1})$ knowledge transfer by KL divergence
    \STATE $G_{t+1}\gets\{f_s^{t+1},c_s^{t+1}\}$
\ENDFOR
\RETURN $G_{T+1}$
\end{algorithmic}
\end{algorithm}

\subsection{Data Augmentation for Server Dataset}
\subsubsection{Generation of Image-Text Samples}
In CLIP-Fed, the server cannot directly utilize client datasets for contrastive learning and knowledge distillation for the purpose of privacy-preserving. 
MLLMs are used to generate and augment server-side datasets without accessing client data~\cite{DBLP:conf/acl/LeeWKMSAMKG24,DBLP:journals/corr/abs-2504-06256}.
Dissimilar to other existing methods that rely on geometric transformation or noise injection, our scheme semantically expands candidate labels to guide image generation. 
For instance, combining object descriptors with contextual modifiers, e.g., ``a small airplane beneath a cloudy sky'' or ``a frog partially obscured by leaves'', naturally introduces variations in scale, occlusion, and background complexity, thereby inherently influencing visual perception.
The synthesized images are then adaptively resized and normalized to match the statistical distribution of the target dataset, thus the alignment with the original feature space is guaranteed.

\subsubsection{Frequency Sensitivity-guided Image Augmentation}
To enhance the robustness of backdoor purification against triggers, we add perturbations to image samples of the server dataset to simulate images with triggers while preserving label information.
Specifically, we first add visible triggers to a clean image $x$ to obtain $\bar{x}$, then apply the Discrete Cosine Transform (DCT) to both $x$ and $\bar{x}$ and divide their spectra into multiple frequency bands.
By comparing the similarities between the images $x$ and $\bar{x}$ in different domains, we identify the frequency intervals that are the most sensitive to triggers. 
We assume that frequency intervals with the lowest similarity exhibit the highest sensitivity to triggers.
The DCT spectrogram displays low-frequency coefficients clustered in the upper-left and high-frequency components in the lower-right, exhibiting symmetric about the main diagonal.
We segment the spectrogram into discrete frequency bands ordered from low to high, aligned orthogonally to the diagonal.
Specifically, the matrix is unevenly divided into $K$ segments along the diagonal, with the distance from the $k$-th division point to the top-left corner set to $1/2^{(K-k)}$ of the diagonal length.
Then, $K$ frequency bands are extended into $K$ independent spectrograms, represented as $\left(F_1(u, v), \ldots, F_K(u, v)\right)$.
We evaluate the distance between the spectrogram coefficients in each frequency segment using Mean Squared Error (MSE).
Frequency regions with higher MSE values demonstrate lower similarity, indicating higher sensitivity to pixel-block triggers.
The frequency band with the highest MSE is denoted as $F_{high}(u,v)$, where Gaussian noise perturbations are injected into the original image.
The image is then reconstructed into the spatial domain via IDCT, resulting in an enhanced server dataset image with noise added only in the sensitive $F_{high}$ region.
Perturbing only the most sensitive frequency bands better simulates backdoor triggers while preserving model utility.



\subsection{Dynamic Model Filtering
by Clustering}
Clustering presents a potential solution for distinguishing between malicious and benign models in FL. 
In CLIP-Fed, we adopt HDBSCAN for pre-aggregation filtering, following advanced methods~\cite{DBLP:conf/uss/NguyenRCYMFMMMZ22}.
Because high-dimensional parameter spaces suffer from the curse of dimensionality, which reduces the reliability of distance metrics like cosine similarity, we use PCA to project model parameters into a lower-dimensional space where distance computations are more meaningful and robust.
Following the threat assumptions and parameter settings established in Flame~\cite{DBLP:conf/uss/NguyenRCYMFMMMZ22} and FreqFed~\cite{DBLP:conf/ndss/FereidooniPRDS24}, we operate under the premise that benign models constitute the majority of clients while malicious models remain in the minority. 
Accordingly, we set the minimum cluster size to encompass at least 50\% of clients, ensuring the resulting clusters contain predominantly benign model updates. 
While this clustering process enables CLIP-Fed to eliminate most malicious models prior to aggregation, it is acknowledged that the identified benign clusters may still contain some residual malicious models.
After model clustering, CLIP-Fed performs global model aggregation as specified in Eq. (\ref{eq:agg}):
\begin{equation}\label{eq:agg}
    \theta_s^{t+1} = \sum_{i\in S_t} \frac{D_i}{D_{S_t}} \theta_{i}^t, \ \theta \in \{f,c\},
\end{equation}
where $S_t$ denotes the set of clients in the clustering cluster.
In each training round, CLIP-Fed dynamically selects different clients to participate in aggregation, thereby resisting attacks from adaptive adversaries who randomly compromise clients during each round.
The framework subsequently employs post-aggregation backdoor elimination strategies to purify further any remaining backdoor patterns that may persist in the global model.

\begin{table*}[t]
	\centering
    \caption{Performance comparison between CLIP-Fed and other defense methods}
	\begin{tabular}{cccccccc}
		\bottomrule
		 \multirow{2}*{Datasets} & \multirow{2}*{Attack}&FedAvg & FLTrust &FLAME
        &FoolsGold
        &FEDCPA
        &CLIP-Fed\\
        \cline{3-8}
       & & MA/ASR(\%)&MA/ASR(\%)&MA/ASR(\%)&MA/ASR(\%)&MA/ASR(\%)&MA/ASR(\%)\\
        \hline

         \multirow{4}*{CIFAR-10}
        & Badnets & 78.87 / 96.11 &73.53 / 84.78 &70.75 / 3.55 &72.66 / 83.87 &78.65 / 38.37 &81.50 / \textbf{2.77}  \\
         & DBA & 78.23 / 91.69 &76.24 / 46.69 &72.75 / 4.22 &77.83 / 56.29 &77.57 / 35.05 &81.75 / \textbf{1.86}  \\
         & LayerAttack & 79.30 / 14.18 &76.30 / 1.42 &72.50 / 2.54 &67.39 / 8.38 &78.39 / 1.77 &76.70 / \textbf{0.19}  \\
         & AFA & 77.43 / 18.05 &76.42 / 6.96 &73.18 / 4.05 &70.77 / 14.03 &76.23 / 4.00 &80.91 / \textbf{1.42}  \\
        \cline{1-8}
        \multirow{4}*{CIFAR-10-LT}
        & Badnets & 70.88 / 80.17 &63.83 / 64.05 &69.13 / 5.16 &66.46 / 40.27 &71.96 / 4.16 &73.07 / \textbf{4.01}  \\
         & DBA & 70.83 / 62.00 &62.22 / 40.22 &69.21 / 4.62 &65.24 / 66.02 &72.36 / 3.47 &72.73 / \textbf{3.38}  \\
         & LayerAttack & 72.30 / 14.18 &68.06 / 8.73 &68.69 / 7.58 &66.15 / 15.38 &71.17 / 3.82 &70.00 / \textbf{0.29}  \\
         & AFA & 71.67 / 17.18 &67.45 / 5.19 &67.34 / 4.34 &47.95 / 7.29 &72.43 / 3.86 &74.05 / \textbf{2.25}  \\
        \cline{1-8}
         \multirow{4}*{CIFAR-100}
        & Badnets & 43.06 / 97.12 &36.07 / 48.48 &45.09 / 38.02 &37.45 / 88.75 &47.68 / 1.19 &45.47 / \textbf{0.39}  \\
         & DBA & 45.96 / 93.68 &41.38 / 56.88 &45.24 / 0.64 &44.90 / 95.15 &48.29 / 1.19 &46.17 / \textbf{0.39}  \\
         & LayerAttack & 46.40 / 10.48 &43.63 / 0.98 &41.59 / 1.74 &47.68 / 2.40 &49.48 / 0.52 &44.94 / \textbf{0.00}  \\
         & AFA & 41.82 / 13.51 &42.64 / 0.51 &44.91 / 0.62 &44.01 / 2.64 &47.61 / 0.59 &44.57 / \textbf{0.30}  \\
       \toprule
       
	\end{tabular}
 \label{tab:Performance comparison}
\end{table*}

\subsection{Feature Rectification via Prototype Alignment}
The proposed approach systematically eliminates backdoor-induced deviations in feature representations while maintaining the model's core classification performance.
In general, backdoor attacks involve malicious clients injecting trigger-infused samples with target labels, causing the model to misclassify these inputs. 
This leads to a shift in the target class prototype, either pulling it toward the attack samples or creating a fake prototype, along with increased intra-class variance.
In this phase, CLIP-Fed utilizes the CLIP model, which is equipped with rich prior knowledge, as an auxiliary module to guide the rectification and retraining of the feature extraction module of global model on the server. 
In backdoor models, the feature extraction module exhibits the increased sensitivity to trigger-associated patterns. 
Existing research~\cite{DBLP:conf/aaai/ShiZYLXQ24} has evidenced that using CLIP's strongly supervised feature prototypes can guide the generation of federated features. 
Inspired by this fact, we employ contrastive learning between CLIP-derived class prototypes and features generated by the global model feature extractor, thereby mitigating prototype deviations caused by backdoor trigger interference.
As shown in Eq.~\ref{eq:lossproto1}, in the first part, we treat the output of the feature extractor $v_i^c$ (Sample $i$ belonging to class $c$) and the output of CLIP's visual encoder $e_{c, \text{clip}}^I$ as positive pairs, while considering the output of the feature extractor $v_i^c$ and other class prototypes $e_{k, \text{extractor}}$ from the feature extractor as negative pairs.
\begin{equation}\label{eq:lossproto1}
\mathcal{L}_i^{\text{proto-1}}=-\log \frac{\exp \left(sim\langle v_i^c,e_{c, \text{clip}}^I\rangle / \tau\right)}{\sum_{k=1[k \neq c]}^{C}\exp \left(sim\langle v_i^c, e_{k, \text{extractor}}\rangle / \tau\right)},
\end{equation}
where $sim\left\langle \cdot \right\rangle$ denotes cosine similarity and $\tau$ is a temperature hyperparameter.
As shown in Eq.~(\ref{eq:lossproto2}), in the second part, we form positive pairs between the feature extractor's output $v_i^c$ and outputs of the CLIP's text encoder $e_{c, \text{clip}}^T$, with negative pairs being the feature extractor's output $v_i^c$ and other class prototypes $e_{k, \text{clip}}^T$ from the CLIP's text encoder.
\begin{equation}\label{eq:lossproto2}
\mathcal{L}_i^{\text{proto-2}}=-\log \frac{\exp \left(sim\langle v_i^c,e_{c, \text{clip}}^T\rangle / \tau\right)}{\sum_{k=1[k \neq c]}^{C}\exp \left(sim\langle v_i^c, e_{k, \text{clip}}^T\rangle / \tau\right)}.
\end{equation}
The total prototype contrastive loss as shown in Eq. (\ref{eq:retrain}): 
\begin{equation}\label{eq:retrain}
\mathcal{L}_{\text {pcl}}=\sum_{i=1}^{C\times m}\left(\mathcal{L}_i^{\text{proto-1}}+\mathcal{L}_i^{\text{proto-2}}\right).
\end{equation}
%
\subsection{Global Model Knowledge Transfer}
To thoroughly eliminate residual backdoor patterns in the global model, CLIP-Fed employs knowledge transfer to purify the global model by using CLIP's prior knowledge.
CLIP acts as a ``teacher'' model and the global model serves as the ``student'' model.
During each iteration, CLIP provides guidance to the global model using the server dataset, progressively reducing the impact of the backdoor.
We utilize the KL divergence to measure the discrepancy between logits distributions of the global model and CLIP on the annotated dataset. 
Specifically, the loss function for the knowledge transfer process is formulated as
\begin{equation}\label{eq:kd}
\mathcal{L}_{\text{kt}}=\mathcal{L}_{c e}\left(y, l_{\text{global}}\right)+\beta \cdot \mathcal{D}_{\text{KL}}\left(l_{\text{clip}} \| l_{\text{global}}\right),
\end{equation}
where $l_{\text{global}}$ and $l_{\text{clip}}$ denote the output logits of the global model and CLIP.
This loss of distillation ensures that the global model aligns its predictions with CLIP's robust, backdoor-free representations, thus eradicating malicious patterns while preserving model utility.

\section{Experiments}

\begin{table}[t]
	\centering
    \caption{Performance comparison between CLIP-Fed w/o FR, CLIP-Fed w/o KD, and CLIP-Fed}
	\begin{tabular}{ccccc}
		\bottomrule
		 \multirow{2}*{Datasets} & \multirow{2}*{Attack} &w/o FR&w/o KD&CLIP-Fed
        \\
        \cline{3-5}
      & & MA/ASR&MA/ASR&MA/ASR\\
        \hline
       \multirow{4}*{CIFAR-10}
        & BadNets & 75.88/6.30 &75.08/7.37&\textbf{81.43}/\textbf{2.62} \\
        & DBA & 76.57/3.31 &75.73/4.01&\textbf{81.56}/\textbf{2.05} \\
        & LayerAttack &75.43/6.15 &73.52/5.39&\textbf{77.10}/\textbf{0.19} \\
        & AFA & 76.17/3.52 &73.79/4.53&\textbf{80.94}/\textbf{1.30} \\
        \cline{1-5}
       \multirow{4}*{CIFAR-10-LT}
        & BadNets & 70.88/8.87 &71.31/6.93&\textbf{73.40}/\textbf{4.11} \\
        & DBA & 70.32/4.30 &70.76/5.66&\textbf{73.00}/\textbf{3.29} \\
        & LayerAttack & 70.10/4.20 &69.55/2.15&\textbf{70.20}/\textbf{0.38} \\
        & AFA & 70.87/4.48 &71.55/3.41&\textbf{73.86}/\textbf{2.16} \\
        \cline{1-5}
        \multirow{4}*{CIFAR-100}
        & BadNets & 44.68/0.63 &44.83/0.45&\textbf{45.56}/\textbf{0.41} \\
        & DBA & 44.35/0.63 &45.44/0.45&\textbf{45.79}/\textbf{0.31} \\
        & LayerAttack & 43.97/0.19 &41.21/0.00&\textbf{44.51}/\textbf{0.00} \\
        & AFA & 45.50/0.57 &44.99/0.34&\textbf{45.61}/\textbf{0.22} \\
       \toprule
       
	\end{tabular}

 \label{tab:Ablation}
\end{table}

\begin{figure*}[t]
  \centering
    \subfloat[CLIP]
  {\includegraphics[width=0.23\textwidth]{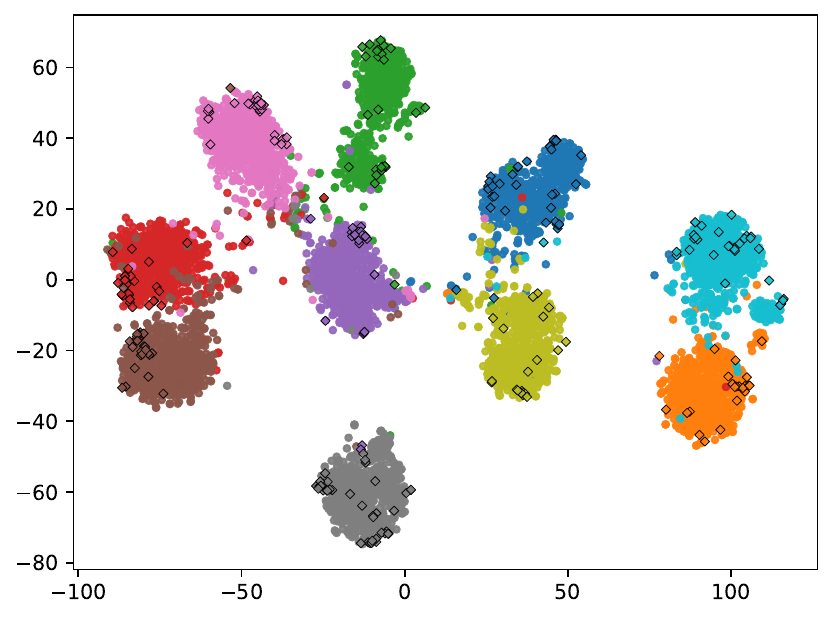}\label{fig:Visualization_CLIP}}
\hfil
  \subfloat[BadNets without defense]
  {\includegraphics[width=0.23\textwidth]{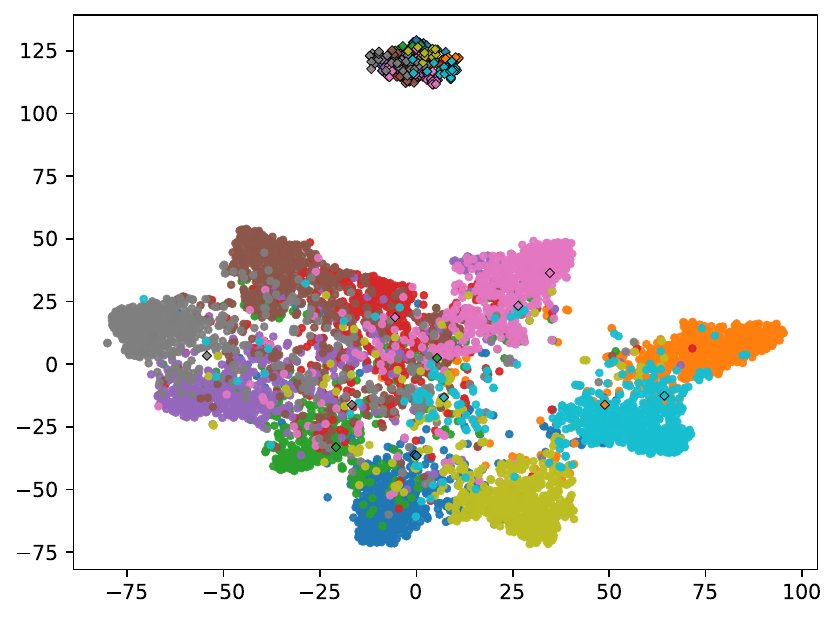}\label{fig:Visualization_nodefense}}
  \hfil
  \subfloat[BadNets with CLIP-Fed 50 epochs]
  {\includegraphics[width=0.23\textwidth]{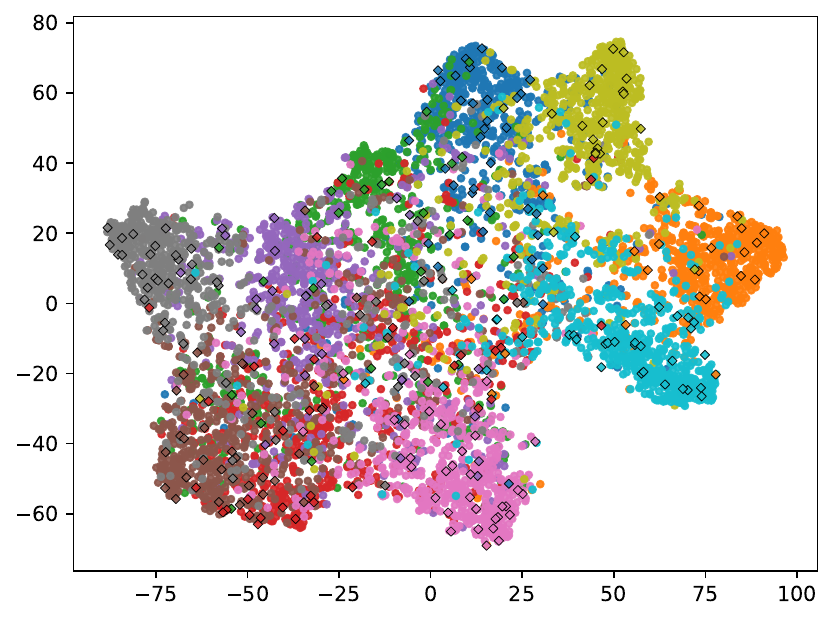}\label{fig:Visualization_50}}
  \hfil
  \subfloat[BadNets with CLIP-Fed 200 epochs]
  {\includegraphics[width=0.23\textwidth]{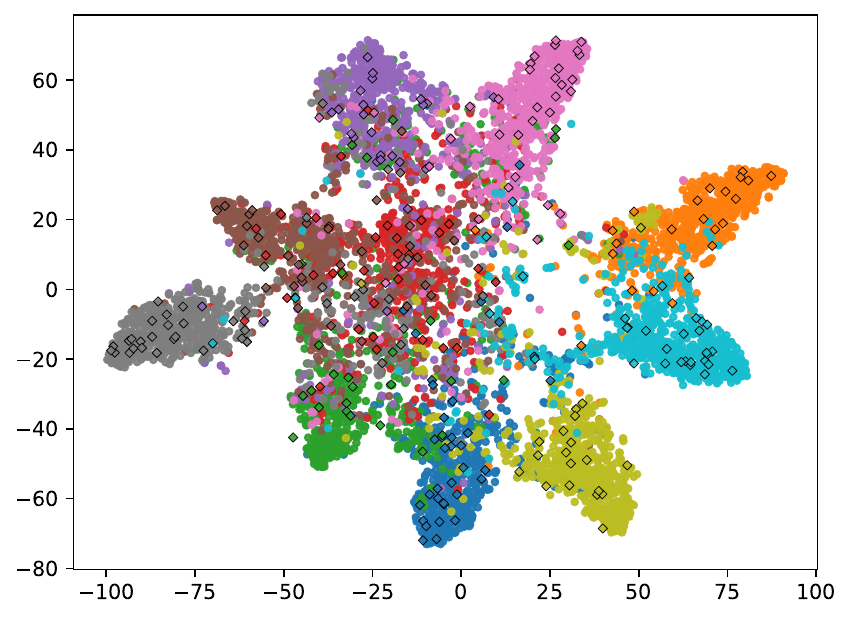}\label{fig:Visualization_200}}
  
  \caption{The visualization of the feature representations under different attacks on CIFAR-10-LT with/without CLIP-Fed.}
  \label{fig:Visualization}
\end{figure*}

\begin{figure}[t]
  \centering

  \subfloat[Original Image]
  {\includegraphics[width=0.3\columnwidth]{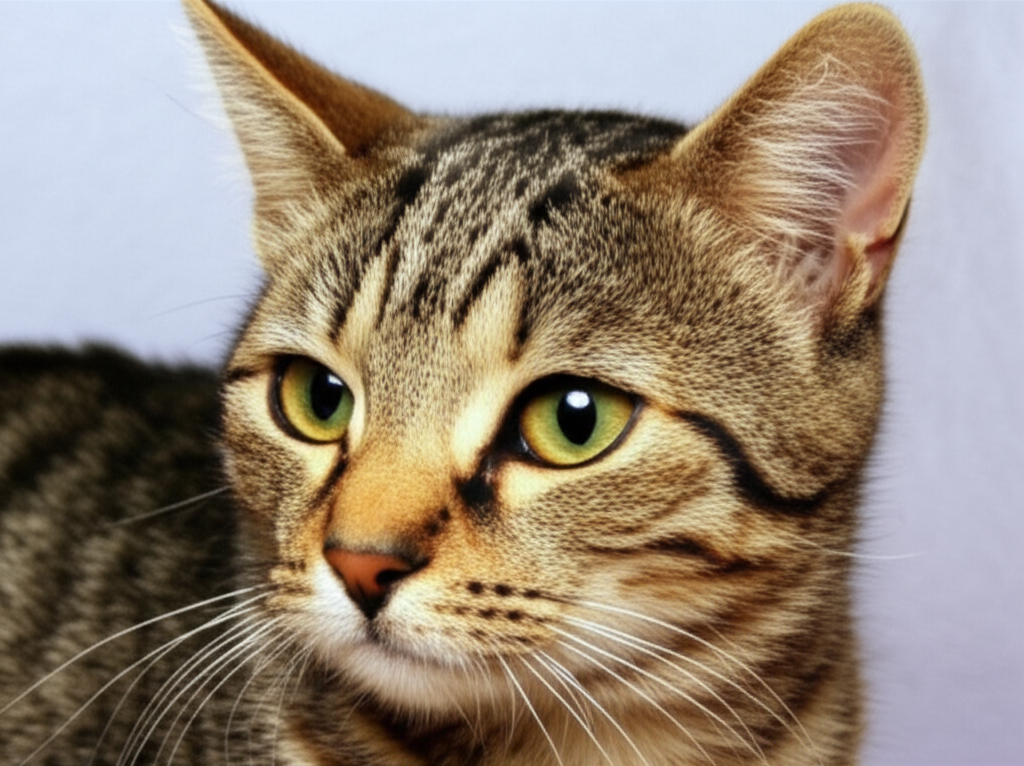}\label{fig:Frequency_raw}}
  \hfil
  \subfloat[Image with Trigger]
{\includegraphics[width=0.3\columnwidth]{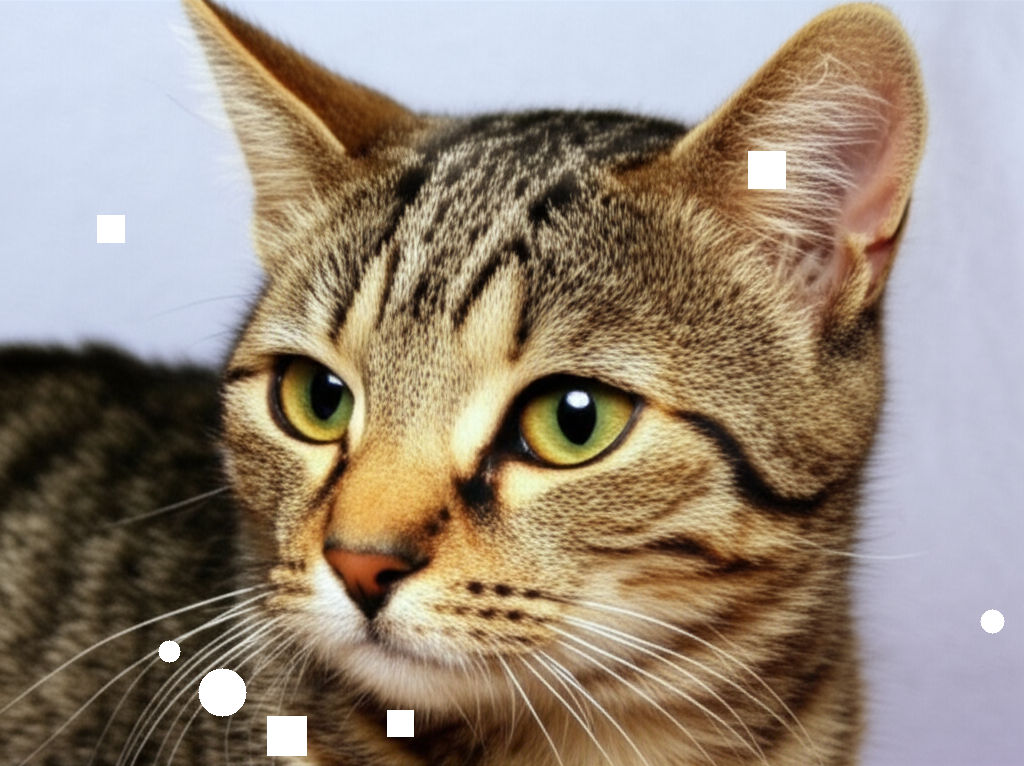}\label{fig:Frequency_trigger}}
  \hfil
  \subfloat[Frequency-perturbed Image]
{\includegraphics[width=0.3\columnwidth]{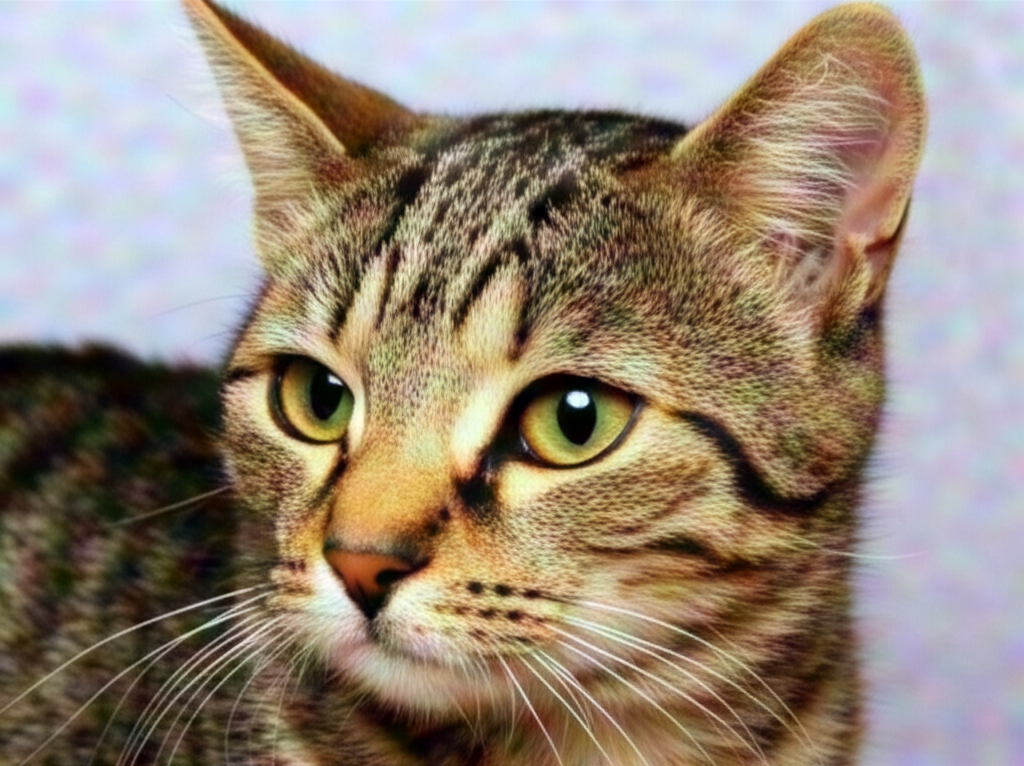}\label{fig:Frequency_noise}}

  \caption{Comparison of original, triggered, and frequency-perturbed images.}
  \label{fig:Frequency}
\end{figure}

\begin{figure}[t]
\centering
\includegraphics[width=0.7\columnwidth]{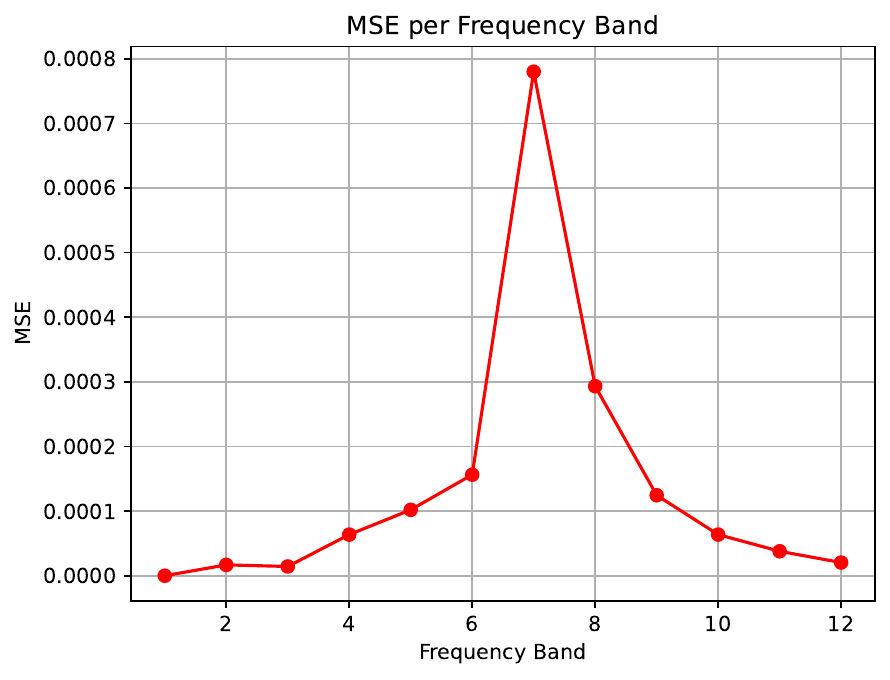} 
  
  \caption{Comparison of MSE values between the original image and the image with pixel blocks in different frequency domain intervals.}
  \label{fig:Frequency_MSE}
\end{figure}

\begin{figure*}[t]
  \centering
    \subfloat[MA with 20\% malicious clients]
  {\includegraphics[width=0.33\textwidth]{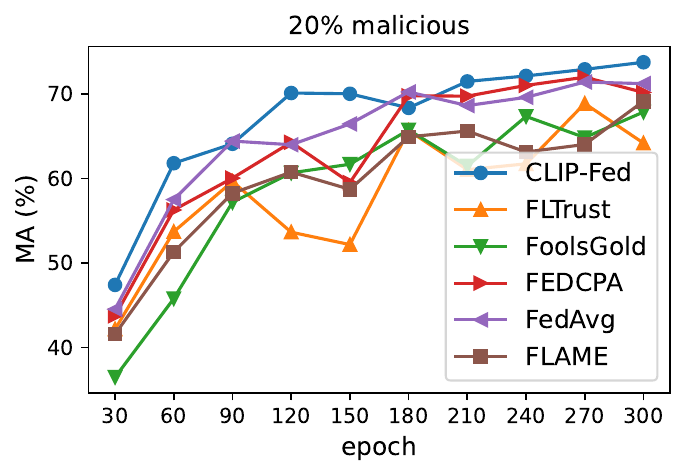}\label{fig:ACC_20}}
\hfil
  \subfloat[MA with 30\% malicious clients]
  {\includegraphics[width=0.33\textwidth]{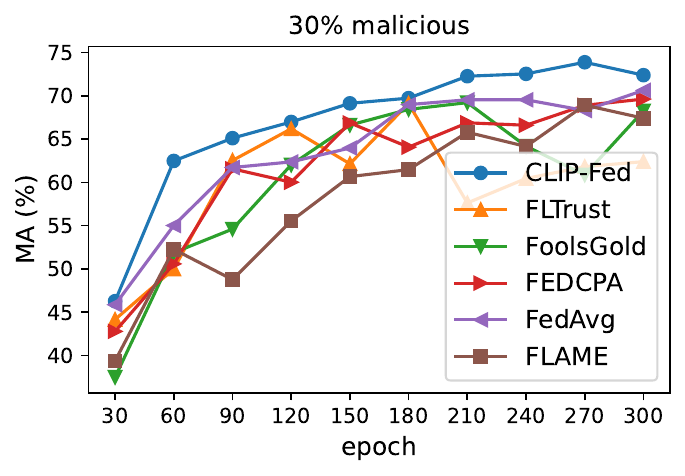}\label{fig:ACC_30}}
  \hfil
  \subfloat[MA with 40\% malicious clients]
  {\includegraphics[width=0.33\textwidth]{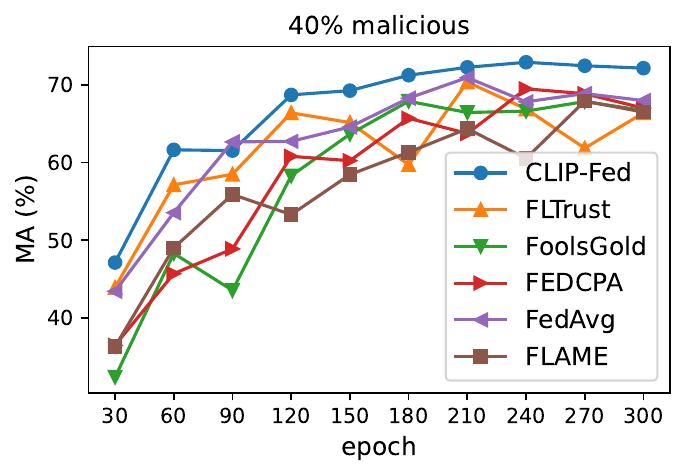}\label{fig:ACC_40}}
\hfil
  \subfloat[ASR with 20\% malicious clients]
  {\includegraphics[width=0.33\textwidth]{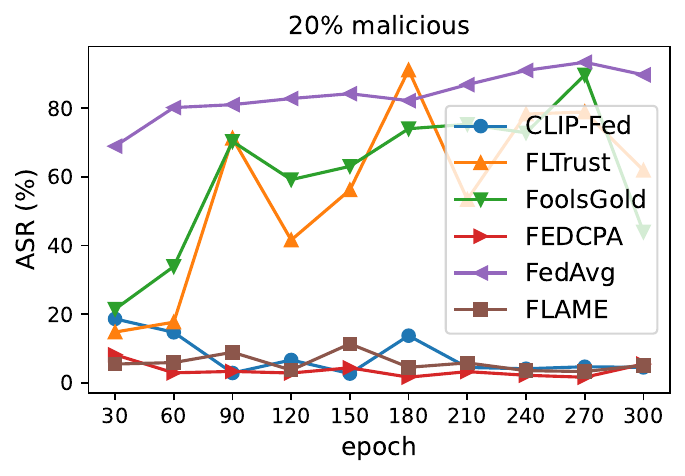}\label{fig:BA_20}}
  \hfil
  \subfloat[ASR with 30\% malicious clients]
  {\includegraphics[width=0.33\textwidth]{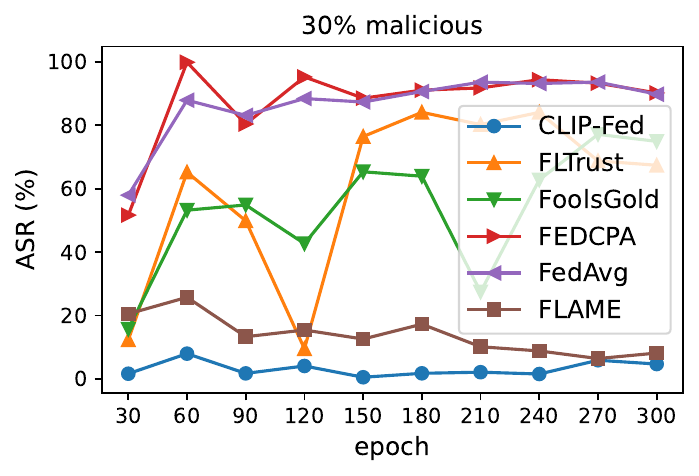}\label{fig:BA_30}}
  \hfil
  \subfloat[ASR with 40\% malicious clients]
  {\includegraphics[width=0.33\textwidth]{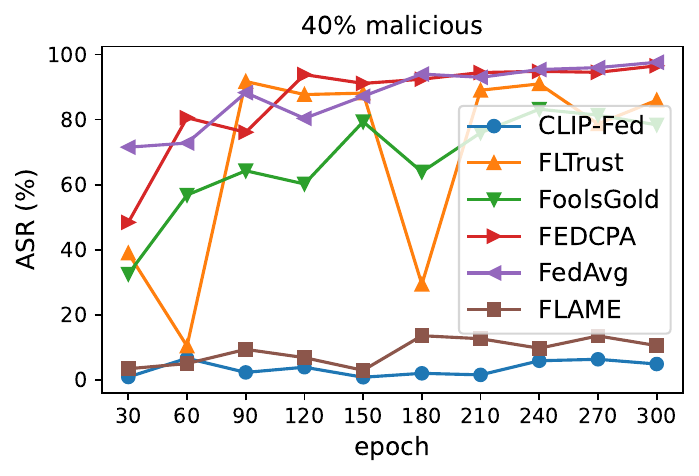}\label{fig:BA_40}}

  \caption{Comparison of MA and ASR against BadNets attacks with varying malicious client proportions under CIFAR-10-LT.}
  \label{fig:Number of Malicious Clients}
\end{figure*}

\begin{table*}[t]
	\centering
    \caption{Performance comparison against BadNets attacks with different proportions of malicious clients under CIFAR-10-LT}
	\begin{tabular}{ccccccc}
		\bottomrule
		 \multirow{2}*{\makecell[c]{Proportions of\\ Malicious Clients}}&FedAvg & FLTrust &FLAME
        &FoolsGold
        &FEDCPA
        &CLIP-Fed\\
        \cline{2-7}
       & MA/ASR(\%)&MA/ASR(\%)&MA/ASR(\%)&MA/ASR(\%)&MA/ASR(\%)&MA/ASR(\%)\\
        \hline

         20\% & 71.21 / 89.74 &64.17 / 61.93 &69.13 / 5.16 &67.85 / 43.99 &70.17 / 5.52 &73.74 / 4.54  \\
         30\% & 70.62 / 89.81 &62.39 / 67.40 &67.40 / 8.17 &68.25 / 74.94 &69.61 / 90.22 &72.37 / 4.70  \\
         40\% & 67.98 / 97.63 &66.40 / 85.97 &66.49 / 10.59 &66.71 / 78.41 &67.01 / 96.54 &72.15 / 4.93  \\
       \toprule
       
	\end{tabular}
 \label{tab:Number of Malicious Clients}
\end{table*}

\begin{table}[t]
	\centering
    \caption{Performance of CLIP-Fed against BadNets attacks with different degrees of Non-IID}
	\begin{tabular}{ccccc}
		\bottomrule
		 \multirow{3}*{\makecell[c]{Degree of \\Non-iid}}&\multicolumn{2}{c}{CIFAR-10}&\multicolumn{2}{c}{CIFAR-100}
        \\
        \cline{2-5}
      & No Defense& CLIP-Fed&No Defense&CLIP-Fed\\
      \cline{2-5}
      & MA/ASR& MA/ASR&MA/ASR&MA/ASR\\
        \hline
        0.2& 78.51/97.04 & 82.41/2.21 &45.06/97.12&45.72/0.77 \\

        0.4& 77.25/95.63 & 81.50/2.77 &44.00/94.49&45.68/0.44 \\

        0.6& 77.42/95.86 & 78.23/0.90 &43.57/91.05&45.60/0.32 \\
       \toprule
	\end{tabular}

 \label{tab:Non-IID}
\end{table}

 \begin{table*}[t]
	\centering
    \caption{Defense performance against BadNets attacks with different attack strength under CIFAR-10-LT}
	\begin{tabular}{ccccccc}
		\bottomrule
		Poisoning Ratio & \multicolumn{2}{c}{60\%} & \multicolumn{2}{c}{80\%}& 
        \multicolumn{2}{c}{100\%}\\
        
        \hline
      Defense& No Defense& CLIP-Fed& No Defense& CLIP-Fed&No Defense& 
     CLIP-Fed\\
      
      \hline
      Trigger Size& MA/ASR(\%)&MA/ASR(\%)&MA/ASR(\%)&MA/ASR(\%)&MA/ASR(\%)&MA/ASR(\%) \\
      
        \hline
    1$\times$1&  70.06 / 11.68&71.83 / 3.92&67.34 / 67.71&71.33 / 4.67&71.21 / 89.74&72.49 / 4.16\\
    2$\times$2& 70.45 / 15.43&72.12 / 4.86&69.15 / 77.74&71.69 / 4.07&71.05 / 94.37&72.11 / 5.40\\
    3$\times$3& 69.56 / 20.08&70.93 / 2.68&68.50 / 88.77&72.58 / 2.34&70.60 / 95.55&72.57 / 2.16\\
    4$\times$4& 69.51 / 28.51&72.53 / 3.99&69.42 / 89.72&72.74 / 2.55&70.40 / 96.11&72.95 / 3.67\\
        
       \toprule
       
	\end{tabular}
 \label{tab:attack strength}
\end{table*}

\noindent\textbf{Datasets.}
We evaluated CLIP-Fed on CIFAR-10, CIFAR-10-LT, and CIFAR-100~\cite{krizhevsky2009learning}. 
CIFAR-10-LT was a variant of CIFAR-10 with long-tail distribution characteristics. 
The number of its category samples decreased exponentially, thereby simulating the category imbalance phenomenon in the real world.
To simulate Non-IID data, we adopted the Pathological Non-IID partitioning~\cite{blanchard2017machine}.

\noindent\textbf{Implementation Details.}
We simulated a FL scenario with 100 clients, where a portion is malicious.
Malicious clients conducted four types of backdoor attacks, including BadNets~\cite{gu2017badnets}, DBA~\cite{xie2019dba}, LayerAttack, and AFA, following an ``all-to-one" strategy.
ResNet-18 was used as the model backbone.
All experiments were conducted on an RTX 4090 GPU with Python 3.9, PyTorch 2.2.2, and CUDA 11.8.
Unless stated otherwise, we simulated Non-IID client data distributions and employed the ViT-L/14 version of CLIP.

\noindent\textbf{Attack Approach.}
We evaluated the defense effectiveness of CLIP-Fed against several existing FL backdoor attacks.
Specifically, the adversary adopted the following attack strategies.
(1) BadNets~\cite{gu2017badnets}, a typical backdoor attack method that directly embedded a static trigger into training samples;
(2) Distributed Backdoor Attack (DBA)~\cite{xie2019dba}, a federated backdoor attack that decomposed a global trigger pattern and embedded its fragments into various clients;
(3) LayerAttack, a hierarchical poisoning method targeting critical layers in neural networks, which was particularly effective against distance-based defense mechanisms;
(4) AFA, a variant of LayerAttack that performed hierarchical bit flipping on sensitive layers, effective against sign-based defenses.
In addition, we assumed that the adversary adopted an ``all to one'' attack strategy when a backdoor attack was launched, which meant the adversary added triggers to samples of all categories and changed their labels to a target category.


\noindent\textbf{Baselines and Existing Defenses.}
We compared CLIP-Fed with representative backdoor defense in FL, including
FedAvg~\cite{mcmahan2017communication} (a standard aggregation baseline without defense),
FLAME~\cite{DBLP:conf/uss/NguyenRCYMFMMMZ22} (combining clustering and differential privacy),
FLTrust~\cite{cao2021fltrust} (a server-side validation-based defense method),
FoolsGold~\cite{DBLP:conf/raid/FungYB20} (a defense scheme based on the similarity of contributions among clients), and FEDCPA~\cite{han2023towards} (focusing on critical parameter similarity).

\subsection{Performance Comparison}
Table~\ref{tab:Performance comparison} presented a comparison of Main Task Accuracy (MA) and Attack Success Rate (ASR) across various methods under distinct data distributions and backdoor attacks.
The results demonstrated that CLIP-Fed consistently achieved effective backdoor mitigation in various attacks and datasets.
Specifically, CLIP-Fed had the lowest ASR under all four attack types on both CIFAR-10 and CIFAR-10-LT datasets.
On CIFAR-10, our scheme achieved ASR reductions of 0.78\%, 2.36\%, 2.35\%, and 2.63\% under BadNets, DBA, LayerAttack, and AFA, respectively, compared to SOTA defense FLAME.
On CIFAR-10-LT dataset that had a significant long-tail distribution, our scheme reduced ASR by 0.15\%, 0.09\%, 3.53\%, and 1.61\% under the same four attacks, compared to the SOTA defense FEDCPA.
In terms of MA, our scheme also had advantages over existing solutions, as it achieved backdoor defense without significantly compromising task performance.
This indicated that CLIP-Fed was able to accurately filter out malicious models while preserving contributions of benign ones, performing a higher-level effectiveness than standalone pre-aggregation detection or post-aggregation noise injection methods.
On CIFAR100 dataset with a higher-level complexity of classification tasks, CLIP-Fed achieved the best defense under four types of attacks.

\subsection{Ablation Study}
We conducted comprehensive ablation experiments on CIFAR-10 under backdoor attacks to thoroughly validate the effectiveness of CLIP-Fed's core components.
Two variants of CLIP-Fed were evaluated, which were CLIP-Fed w/o FR (without feature extractor rectification) and CLIP-Fed w/o KD (without knowledge distillation). 
Table~\ref{tab:Ablation} showed comparisons of the accuracy on the global model and the defense performance under identical experiment setups.
We observed that both CLIP-Fed w/o FR and CLIP-Fed w/o KD achieved lower ASR than FedAvg (as shown in Table~\ref{tab:Performance comparison}) across all three datasets and four types of attacks.
The results indicated that individually applying either feature extractor retraining or global model knowledge distillation could partially eliminate the association between trigger patterns and target labels, thereby mitigating backdoor effects.
Thus, CLIP-Fed combined both feature retraining and knowledge distillation, so that it achieved the highest MA and lowest ASR in call cases.

\subsection{Visualization of Backdoor Elimination}
To evaluate the purification effectiveness after multiple rounds of backdoor elimination (feature rectification and global model knowledge transfer), we examined changes in the association between trigger samples and target labels before/after defense. 
As shown in Figure~\ref{fig:Visualization}, t-SNE was used to visualize the feature space distribution of test samples under BadNets attacks, both with and without CLIP-Fed.
We added triggers to 5\% of the test samples.
In visualization, samples containing triggers were outlined in a black color.
Figure~\ref{fig:Visualization_CLIP} presented the visualization of test samples after being processed by the CLIP model. 
The CLIP model demonstrated strong robustness to trigger samples and successfully clustered samples from the same class together in the feature space. 
This suggested that using CLIP's prior knowledge to guide FL in backdoor purification was a feasible approach.
Figure~\ref{fig:Visualization_nodefense} showed the feature space clustering after a backdoor attack without any defense. 
We observed that most samples with triggers formed a separate cluster, indicating a successful backdoor attack.
Figures~\ref{fig:Visualization_50} and \ref{fig:Visualization_200} showed clustering results after training with CLIP-Fed for 50 and 200 epochs, respectively. 
The trigger samples were no longer isolated and rejoined the clusters of clean samples from the same class.
The results indicated that CLIP-Fed disrupted backdoor feature clustering and weakened the association between triggers and target labels. 

\subsection{Analysis of the Frequency Sensitivity-guided Image Augmentation}
To assess the proposed frequency sensitivity-guided image augmentation, we compared the original image, a pixel-block triggered image, and our frequency-perturbed image, as shown in Figure~\ref{fig:Frequency}.
Visual results demonstrated that our scheme avoided introducing visible distortion, thereby preserving the semantic content.
We computed MSE between the DCT spectra of the clean and triggered images across $k$ frequency bands.
As shown in Figure~\ref{fig:Frequency_MSE}, the 7th frequency band exhibited the highest MSE, indicating the strongest sensitivity to trigger artifacts.
We injected Gaussian noise into $F_{high}(u,v)$, and reconstructed the image via IDCT (refer to Figure~\ref{fig:Frequency_noise}).
This approach simulated potential backdoor signals without compromising visual quality. 
Moreover, by focusing perturbations on the most trigger-sensitive frequency regions, the augmented images served as effective surrogates for trigger patterns, enabling the global model to generalize more effectively against diverse backdoor attacks during training.

\subsection{Parameter Impact}

\noindent\textbf{Impact of Number of Malicious Clients.}
Table~\ref{tab:Number of Malicious Clients} and Figure~\ref{fig:Number of Malicious Clients} presented a comparison of the defense performance of CLIP-Fed and existing methods under varying proportions of malicious clients.
When the proportion of malicious clients was 0.2, FEDCPA, FLAME, and CLIP-Fed demonstrated comparable effectiveness in reducing ASR.
However, as the proportion exceeded 0.2, the defense performance of existing methods began to degrade to varying extents.
In particular, FEDCPA experienced the most significant drop in performance, with an ASR of 90.22\% when the malicious client ratio was 0.3, and 96.54\% when it increased to 0.4.
In contrast, CLIP-Fed consistently maintained strong defense performance across all settings.
When the proportion of malicious clients was 0.3 and 0.4, the ASR of CLIP-Fed remained significantly lower than that of the SOTA method FLAME by 3.47\% and 5.66\%, respectively.
In terms of MA, CLIP-Fed also performed favorably, surpassing FedAvg by 2.53\%, 1.75\%, and 4.17\% at malicious client ratios of 0.2, 0.3, and 0.4, respectively.
As shown in Figure~\ref{fig:Number of Malicious Clients}, CLIP-Fed consistently demonstrated a clear advantage across all stages of training.
A single pre-aggregation defense strategy made existing methods vulnerable to high proportions of malicious clients, whereas CLIP-Fed's post-aggregation backdoor purification supplements the defense capability under such conditions.
Overall, compared with existing approaches, CLIP-Fed effectively defended against backdoor attacks across different malicious client ratios while minimizing the degradation of task performance in FL.

\noindent\textbf{Impact of Non-IID.}
Table~\ref{tab:Non-IID} presented the defense performance of CLIP-Fed against BadNets attack under varying degrees of Non-IID data distribution.
On CIFAR-10 dataset, the CLIP-Fed approach achieved an average MA of 80.71\% and an average ASR of 1.96\%.
On CIFAR-100 dataset, the average MA was 45.67\% and the average ASR was 0.51\%. 
Across both datasets, the degree of Non-IID distribution caused only minor fluctuations in MA and ASR, and its impact on defense performance was negligible.
The reason for these phenomena was because heterogeneous client data distributions primarily affected the filtering of malicious models before aggregation, while CLIP-Fed's backdoor purification strategy was decoupled from model aggregation. 
The above clearly demonstrated that CLIP-Fed was effective under different levels of Non-IID data distribution, thereby maintaining a high MA and a low ASR for consistent robust backdoor defense.

\noindent\textbf{Defense under Different Attack Strength.}
To investigate CLIP-Fed's defensive performance against backdoor attacks with varying strengths, we evaluate attacks employing different trigger sizes and poisoning ratio on the CIFAR-10-LT dataset.
Trigger size and poisoning ratio served as crucial factors affecting the attack strength.
When attackers became aware of the deployed defense mechanisms, they might avoid defense strategies based on malicious model screening rules, either increasing trigger size and poisoning to enhance attack chance or reducing trigger size and poisoning to improve backdoor concealment. 
As shown in Table~\ref{tab:attack strength}, in the absence of defense, MA gradually declined while ASR increased as the trigger size became larger.
Larger triggers resulted in more distinguishable features, making it easier for the model to associate them with the target label.
Additionally, higher poisoning rates further intensified the attack, resulting in a sharper drop in model performance.
When CLIP-Fed was applied, backdoor were effectively prevented from being embedded into the global model, regardless of trigger sizes or poisoning rates.
The connection between trigger features and the target label was severed, and residual backdoor patterns in the global model were removed, making the defense independent of attack strength.
CLIP-Fed demonstrated consistent effectiveness in mitigating backdoor effects across adaptive attacks with various trigger sizes and poisoning ratios.

\begin{figure}[t]
  \centering

  \subfloat[MA with different version of CLIP]
  {\includegraphics[width=0.22\textwidth]{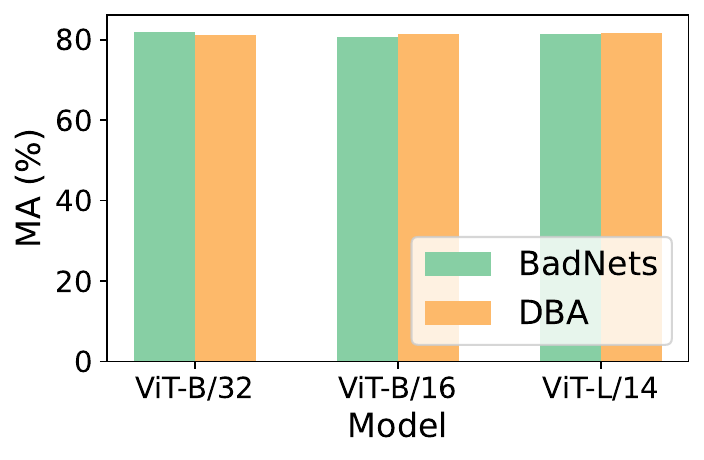}\label{fig:Version of CLIP MA}}
  \hfil
  \subfloat[ASR with different version of CLIP]
  {\includegraphics[width=0.23\textwidth]{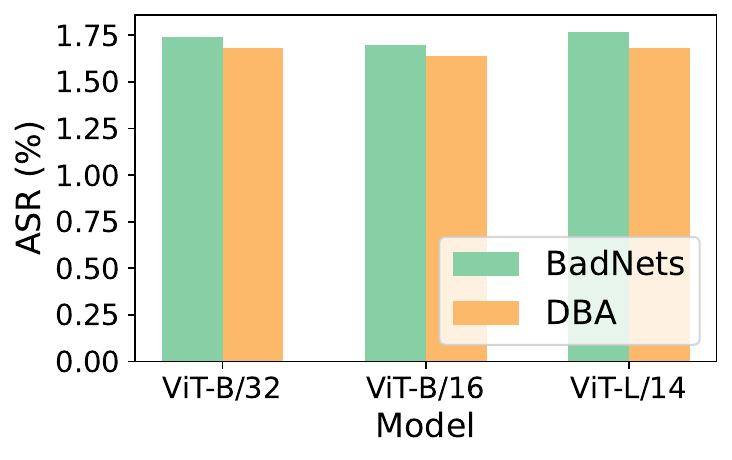}\label{fig:Version of CLIP ASR}}
  
  \caption{Comparison of MA and ASR in resisting BadNets and DBA attacks with different version of CLIP under CIFAR-10.}
  \label{fig:Version of CLIP}
\end{figure}

\noindent\textbf{Impact of Distinct CLIP Versions on Backdoor Defense Performance.}
We evaluated the impact of different versions of CLIP~\cite{radford2021learning} on the backdoor defense performance.
Specifically, we adopted ViT-B/32, ViT-B/16, and ViT-L/14 as VLPs.
Figure~\ref{fig:Version of CLIP} showed the comparison of MA and ASR on CIFAR-10 dataset under BadNets~\cite{gu2017badnets} and DBA~\cite{xie2019dba} attacks when different CLIP versions were applied.
In terms of accuracy, we observed that CLIP-Fed achieved over 80\% MA across all CLIP versions, evidencing that the rich prior knowledge from each version was sufficient for the CIFAR-10 dataset.
Regarding ASR, we found that the success rate under BadNets attacks was higher than that of DBA attacks, due to the stronger impact of centralized attacks.
However, all CLIP versions achieved an ASR below 2\% under both types of attacks, demonstrating that they could all effectively support CLIP-Fed in defending against backdoor threats.
These findings indicate that the defense performance of CLIP-Fed remains consistently strong across different CLIP versions.
This suggested that CLIP-Fed be not highly sensitive to the scale of the VLPs and flexibly adapt to various deployment scenarios without significant degradation in defense effectiveness.

\section{Conclusion}
In this paper, we proposed CLIP-Fed that is a general novel backdoor defense scheme guided by the VLP model CLIP.
Without accessing client data or assuming homogeneous distributions, CLIP-Fed enhanced overall effectiveness by combining pre-aggregation malicious model filtering with post-aggregation purification.
Leveraging MLLMs and frequency analysis, we constructed and enhanced a server dataset, addressing privacy concerns and improving coverage of diverse backdoor triggers with minimal utility loss.
We further proposed a global model purification strategy based on contrastive learning and knowledge distillation, which corrected class prototype deviations and weakened the link between triggers and target labels, thus improving robustness against potential trigger samples.
Extensive experiments on multiple datasets demonstrated that CLIP-Fed outperformed existing methods in defending against backdoor attacks under significantly varying data distributions and malicious proportions of clients.

\bibliographystyle{ACM-Reference-Format}
\bibliography{sample-base}










\end{document}